\documentclass[conference]{IEEEtran}
\IEEEoverridecommandlockouts
\usepackage{cite}
\usepackage{amsmath,amssymb,amsfonts}
\usepackage{algorithmic}
\usepackage{graphicx}
\usepackage{textcomp}
\usepackage{xcolor}
\usepackage{amsmath}
\usepackage{graphicx}
\usepackage{subcaption}
\usepackage{hyperref}
\usepackage{multirow} 
\usepackage{diagbox}
\usepackage{booktabs}
\usepackage{marvosym}

\def\BibTeX{{\rm B\kern-.05em{\sc i\kern-.025em b}\kern-.08em
    T\kern-.1667em\lower.7ex\hbox{E}\kern-.125emX}}
\begin{document}

\title{Consistency Change Detection Framework for Unsupervised Remote Sensing Change Detection}

\author{\IEEEauthorblockN{Yating Liu}
\IEEEauthorblockA{\textit{School of Artificial Intelligence and Data Science} \\
\textit{University of Science and Technology of China}\\
Hefei, China \\
liuyat@mail.ustc.edu.cn}
\and
\IEEEauthorblockN{Yan Lu $^{\ast}$ \thanks{*Corresponding author}}
\IEEEauthorblockA{\textit{Department of Information Engineering} \\
\textit{The Chinese University of Hong Kong}\\
Hong Kong, China \\
yanlu@cuhk.edu.hk}
}

\maketitle

\begin{abstract}
Unsupervised remote sensing change detection aims to monitor and analyze changes from multi-temporal remote sensing images in the same geometric region at different times, without the need for labeled training data. Previous unsupervised methods attempt to achieve style transfer across multi-temporal remote sensing images through reconstruction by a generator network, and then capture the unreconstructable areas as the changed regions. However, it often leads to poor performance due to generator overfitting.
In this paper, we propose a novel Consistency Change Detection Framework (CCDF) to address this challenge. Specifically, we introduce a Cycle Consistency (CC) module to reduce the overfitting issues in the generator-based reconstruction. Additionally, we propose a Semantic Consistency (SC) module to enable detail reconstruction. Extensive experiments demonstrate that our method outperforms other state-of-the-art approaches.
Code is available at \url{https://github.com/cherry0116/CCDF}.
\end{abstract}


\section{Introduction}
\label{intro}

\begin{figure}[t]
\centering
\begin{minipage}[c]{0.146\textwidth}
\centering
\includegraphics[width=\textwidth]{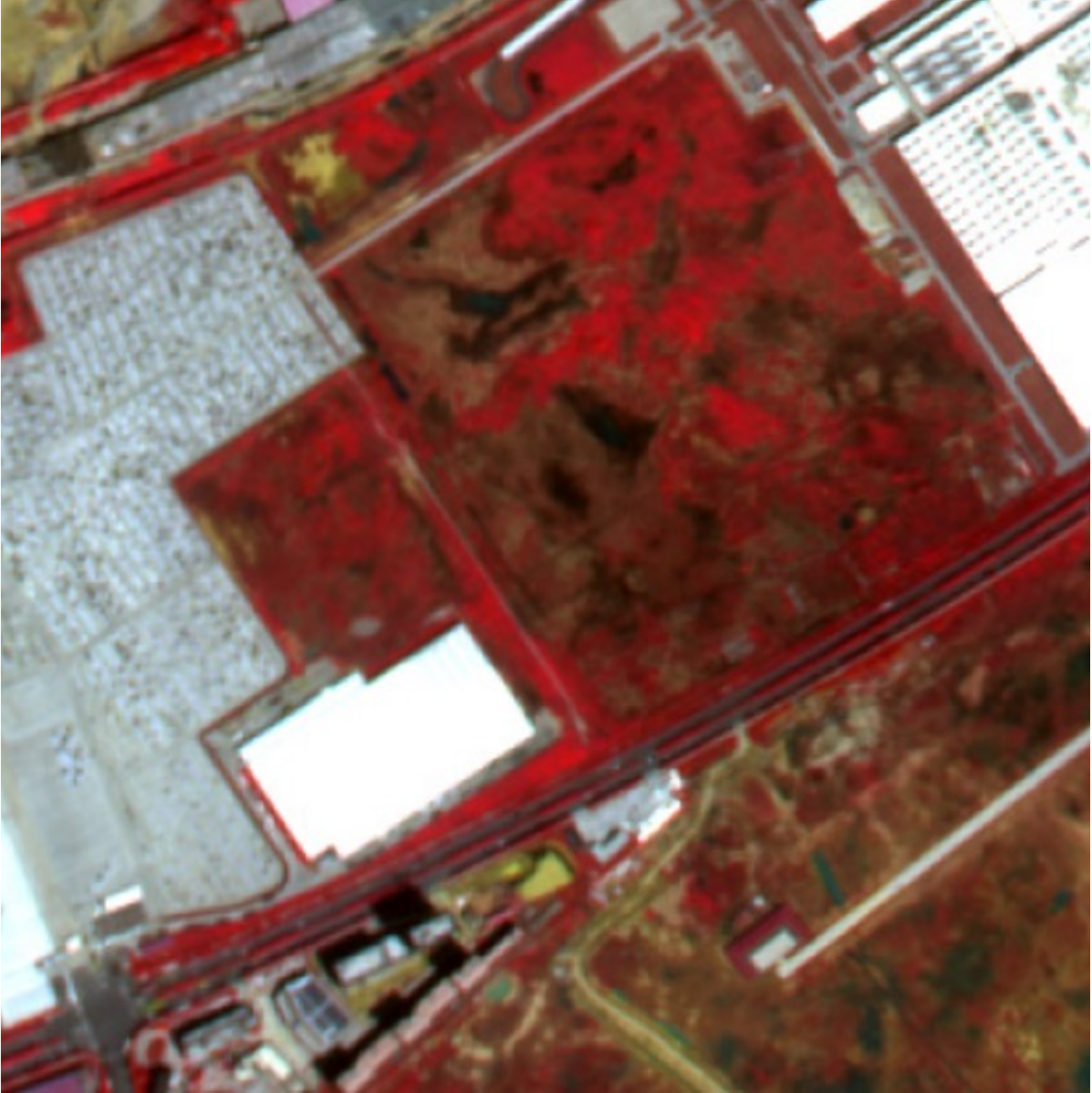}
\subcaption{}
\label{example_t1}
\end{minipage} 
\begin{minipage}[c]{0.146\textwidth}
\centering
\includegraphics[width=\textwidth]{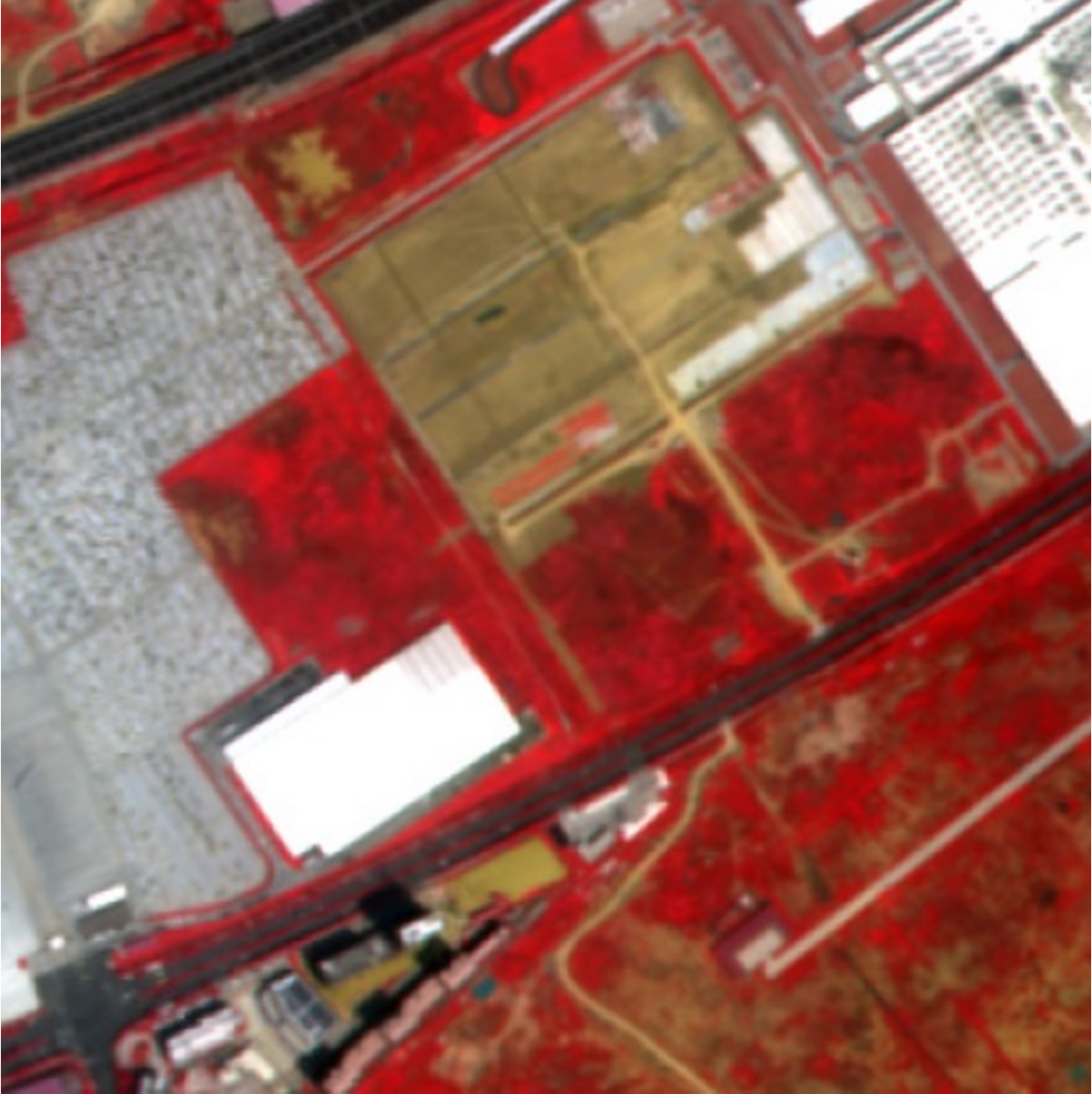}
\subcaption{}
\label{example_t2}
\end{minipage} 
\begin{minipage}[c]{0.146\textwidth}
\centering
\includegraphics[width=\textwidth]{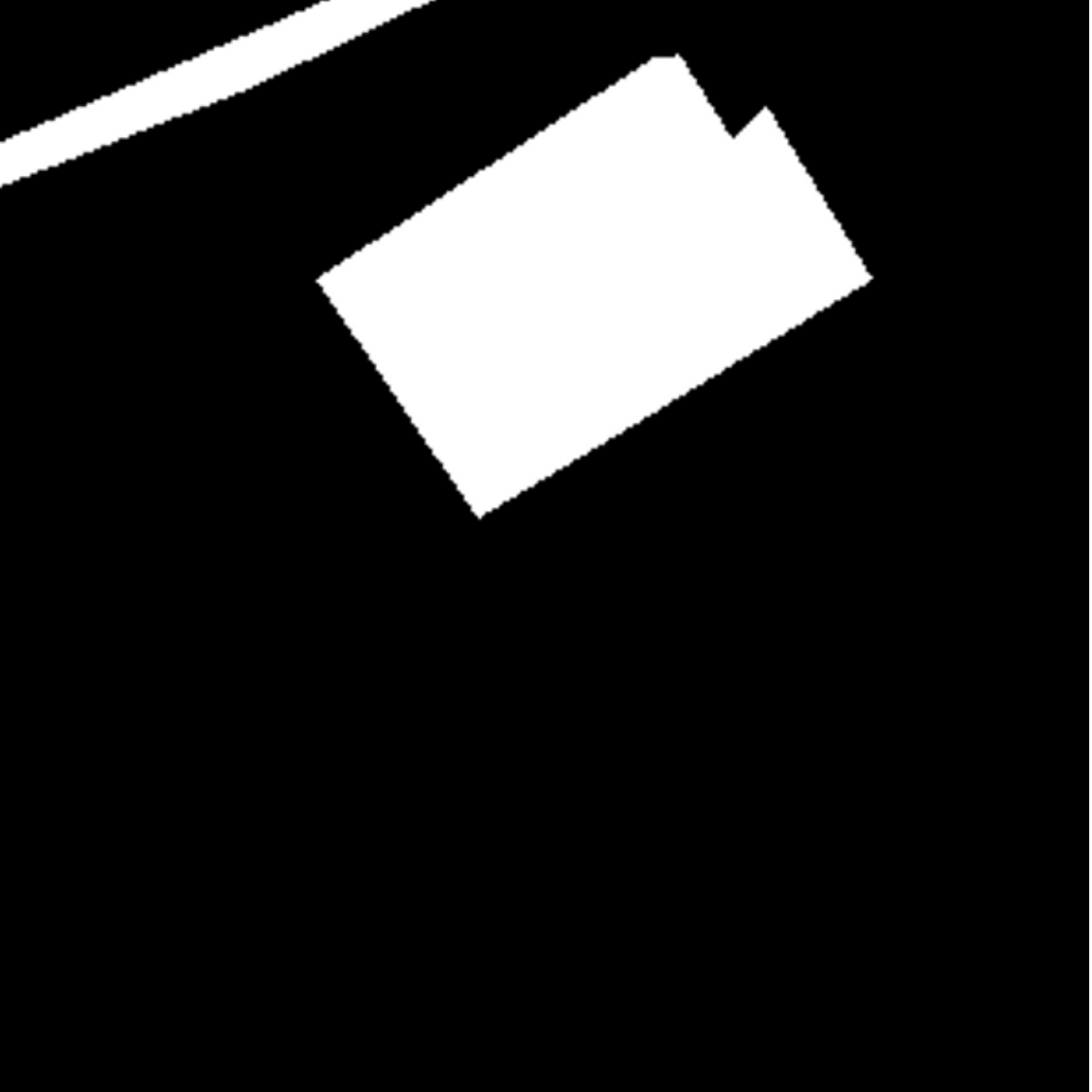}
\subcaption{}
\label{example_ref}
\end{minipage}
\caption{An example of remote sensing change detection. (a) and (b) show remote sensing images of the same geographic region captured at different time points. (c) is the ground truth change map, where the white and black regions represent the changed areas and the unchanged areas, respectively, between (a) and (b).}
\label{example}
\end{figure}

Remote sensing technology has enabled the collection of vast amounts of multi-temporal imagery, providing a rich data source for monitoring and analyzing changes in the Earth's surface and environment \cite{shafique2022deep, zhu2024review}. Detecting and characterizing these changes in the same geographic regions over time is a crucial task in numerous applications, such as urban planning, land cover monitoring, disaster response, and environmental conservation. 
Fig.~\ref{example} presents an example of remote sensing change detection. Fig.~\ref{example}(a) and (b) show remote sensing images of the same geographic region captured at different time points. Fig.~\ref{example}(c) is the ground truth change map, where the white and black regions represent the changed areas and the unchanged areas, respectively, between the images in Fig.~\ref{example}(a) and (b).
Conventional approaches to change detection in remote sensing often rely on supervised learning techniques, which require extensive labeled training data to build robust change detection models \cite{feng2023compact, ma2023stnet, ma2024ddlnet}. However, for large-scale or rapidly evolving environments, acquiring accurate and representative ground truth labels can be labor-intensive and time-consuming. To address this limitation, unsupervised change detection methods have gained increasing attention.

Unsupervised change detection is a challenging task \cite{dong2020self, touati2020anomaly, bergamasco2022unsupervised, chen2021self, lin2024unsupervised}. How to accurately and effectively obtain changes without training labels has become a research difficulty. At present, unsupervised change detection is mainly divided into two branches. One is pre-classification methods \cite{gong2015change, gao2019change, ma2019change}. These approaches generally include two steps: (a) Using a clustering algorithm, such as Fuzzy C-Means (FCM) \cite{chuang2006fuzzy}, to pre-classify the multi-temporal image and select reliable sample image patches. (b) Post-processing methods, such as deep neural networks, are used to obtain the final change map. The main drawback of these methods is that they heavily rely on the accuracy of the pre-classification results, resulting in erroneous transmission or amplification. The other is reconstruction methods \cite{luo2020research, liu2021unsupervised, wu2023fully}. These works aim to achieve style transfer across multi-temporal remote sensing images through generator network reconstruction and identify the unreconstructable areas as the changed regions. Specifically, these methods first decompose the remote sensing image pairs into corresponding patch pairs, and the generator network learns the mapping relationship between the patch pairs at different time points. Due to this patch-level training strategy, the generator network may memorize the specific input-output relationships of the remote sensing patch pairs, rather than learning more generalizable representations \cite{zhang2021understanding, larochelle2009exploring, jacot2018neural}. In the extreme case, the network could completely memorize the correspondences between the input and output patch pairs, effectively eliminating any unreconstructed regions. It will greatly affect the subsequent segmentation of the unreconstructable area, which is not what we want.

In this work, we concentrate on tackling the overfitting problem that arises due to the generator networks' propensity to memorize the specific input-output correspondences.
To this end, we propose a novel Consistency Change Detection Framework (CCDF) to tackle this challenge. Specifically, we introduce a Cycle Consistency (CC) module, which exploits the idea of bidirectional generation. The CC module not only reconstructs from one time point to another but can also reconstruct back, which helps alleviate the problem of overfitting in the generator.
In addition, we propose a Semantic Consistency (SC) module to capture reconstruction details and improve the accuracy of change detection. Extensive experiments demonstrate that our method outperforms other state-of-the-art approaches.  

The contributions of this paper are three-fold.
\begin{itemize}
\item We propose a Cycle Consistency (CC) module to address the overfitting challenges commonly encountered in generator-based reconstruction approaches. The CC module helps the generator network learn more robust and generalizable representations, mitigating the overfitting problem.
\item We present a Semantic Consistency (SC) module to enable a more accurate reconstruction of details in the change detection task. 
\item Extensive experimental evaluations on two benchmark datasets demonstrate that our Consistency Change Detection Framework (CCDF) significantly outperforms other state-of-the-art methods in terms of change detection performance, showcasing its effectiveness and robustness in change detection applications.
\end{itemize}

\begin{figure}[t]
\centering
\includegraphics[width=0.99\linewidth]{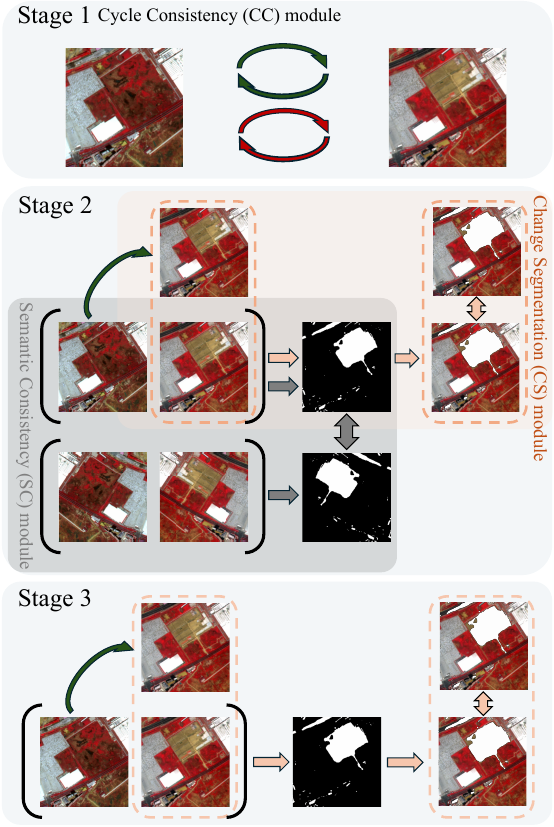} 
\caption{The overview of our method.
\label{pipeline}}
\end{figure}

\section{Proposed Method}
\label{method}

As illustrated in Fig.~\ref{pipeline}, our Consistency Change Detection Framework (CCDF) is trained in three stages.
In the first stage, we leverage generative networks to achieve global style transfer and reconstruction of the bi-temporal remote sensing images. In this stage, we introduce a Cycle Consistency (CC) module to decouple the content and style information, enabling robust reconstruction across time points.
The second stage focuses on learning a segmentation network to capture the regions that cannot be effectively transferred by the generative models. We apply a Change Segmentation (CS) module to capture the unreconstructed regions. Besides, we propose a Semantic Consistency (SC) module to emphasize the reconstruction of details, leading to more accurate segmentation of the changed areas.
Finally, in the third stage, we alternate the fine-tuning of the generative network and the segmentation network, further boosting the overall performance of our change detection framework.

\subsection{Preprocess}

Before inputting the bi-temporal remote sensing image pair ($I_{T_1}$, $I_{T_2}$) into our Consistency Change Detection Framework (CCDF), we apply data standardization to address the issue of scale difference among the image pair:
\begin{small}
\begin{equation}
\begin{aligned}
I_{T_1} = \frac{I_{T_1} - mean(I_{T_1})}{std(I_{T_1})}\\
I_{T_2} = \frac{I_{T_2} - mean(I_{T_2})}{std(I_{T_2})}
\end{aligned}
\label{eq1}
\end{equation}
\end{small}
Here, $mean(\cdot)$ and $std(\cdot)$ represent mean and standard deviation. $I_{T_1}, I_{T_2} \in \mathbb{R}^{W \times H \times C}$, where $W$, $H$ and $C$ are the width, height, and number of channels of a remote sensing image respectively. $T_1$ and $T_2$ represent different acquisition time points of remote sensing images. 

Then we uniformly partition $I_{T_1}, I_{T_2}$ into ($P \times P \times C$)-sized paired patches. $\frac{W}{P} \times \frac{H}{P}$ is the resulting patches number. After that, these paired patches are sent into the CC module to learn global style transfer. To improve readability, we will omit the superscript $p$ and simply write $I_{T_\cdot}$ instead of $I_{T_\cdot}^{p}$ for the patches.

\begin{figure}[t]
\centering
\includegraphics[width=0.99\linewidth]{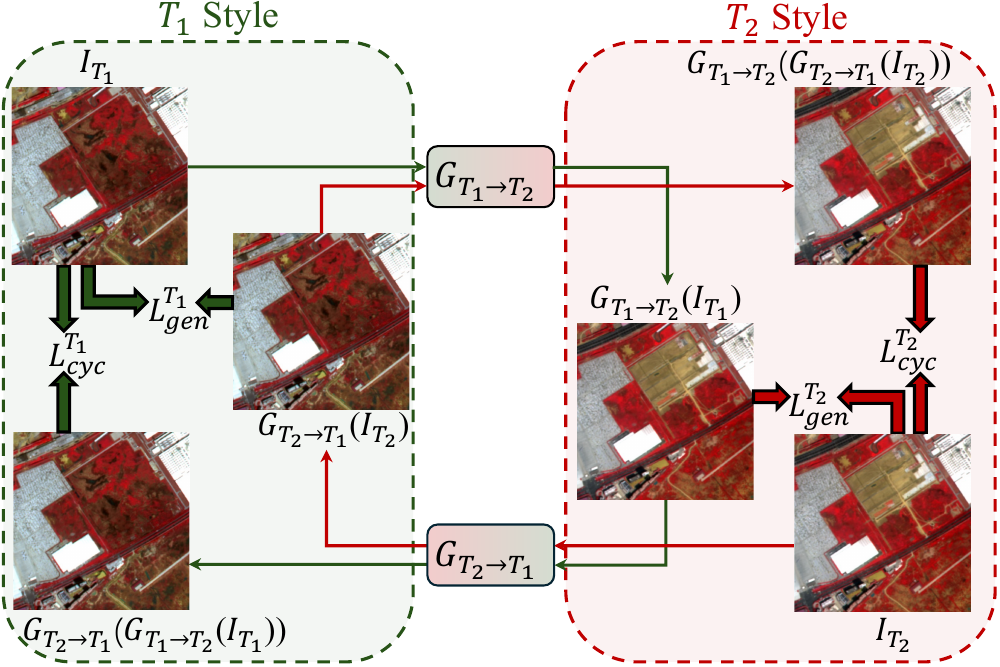} 
\caption{The details of the Cycle Consistency module.
\label{CC}}
\end{figure}

\subsection{Cycle Consistency module}

Inspired by CycleGAN \cite{zhu2017unpaired}, our proposed CC module consists of two generator networks: $G_{T_1 \rightarrow T_2}$ and $G_{T_2 \rightarrow T_1}$. The details of the CC module are shown in Fig.~\ref{CC}. The $G_{T_1 \rightarrow T_2}$ model is tasked with transforming $T_1$ style to $T_2$ style. Conversely, the $G_{T_2 \rightarrow T_1}$ model is tasked with transforming $T_2$ style to $T_1$ style. 
The objective of $G_{T_1 \rightarrow T_2}$ can be expressed as:
\begin{small}
\begin{equation}
G_{T_1 \rightarrow T_2}(I_{T_1}) \approx I_{T_2}
\label{eq2}
\end{equation}
\end{small}
where $\approx$ indicates that $I_{T_1}$ achieves a transfer of image style through $G_{T_1 \rightarrow T_2}$, but the content of $I_{T_1}$ remains unchanged, i.e., $G_{T_1 \rightarrow T_2}(I_{T_1})$ = ($T_1$ content + $T_2$ style). Therefore, $G_{T_1 \rightarrow T_2}(I_{T_1})$ and $I_{T_2}$ are not completely equivalent.
The generator network $G_{T_1 \rightarrow T_2}$ may have a tendency to memorize the specific input-output relationships of the remote sensing patch pairs, rather than learning more generalizable representations. 
Consequently, style transfer can easily become a combination of content and style transfer. To address this, we introduce $G_{T_2 \rightarrow T_1}$ to change the style from $T_2$ back to $T_1$. The goal of cyclic image-to-image style translation can be stated as: 
\begin{small}
\begin{equation}
G_{T_2 \rightarrow T_1}(G_{T_1 \rightarrow T_2}(I_{T_1})) = I_{T_1}
\label{eq3}
\end{equation}
\end{small}
To better fulfill the objective expressed in equation \ref{eq3}, $G_{T_1 \rightarrow T_2}$ aims to preserve the content of $I_{T_1}$ as much as possible, enabling $G_{T_2 \rightarrow T_1}$ to accurately reconstruct the original image $I_{T_1}$.
Therefore, our CC module is able to effectively decouple content and style, thereby achieving global style transfer.

The loss function for $G_{T_1 \rightarrow T_2}$ is calculated as follows:
\begin{small}
\begin{equation}
\begin{split}
\begin{aligned}
&L_{gen}^{T_2} = L_{l_1}(G_{T_1 \rightarrow T_2}(I_{T_1}), I_{T_2}) + \lambda_{cont} \cdot L_{cont}(G_{T_1 \rightarrow T_2}(I_{T_1}), I_{T_2})\\
&L_{l_1}(G_{T_1 \rightarrow T_2}(I_{T_1}), I_{T_2}) = \sum|I_{T_2}-G_{T_1 \rightarrow T_2}(I_{T_1})|\\
&L_{cont}(G_{T_1 \rightarrow T_2}(I_{T_1}), I_{T_2}) = \sum(\phi_{vgg}(I_{T_2}) - \phi_{vgg}(G_{T_1 \rightarrow T_2}(I_{T_1})))^2
\end{aligned}
\end{split}
\label{eq4}
\end{equation}
\end{small}
$L_{l_1}(\cdot, \cdot)$ indicates an L1 loss to reduce the overall difference between $(\cdot, \cdot)$. $L_{cont}(\cdot, \cdot)$ \cite{ledig2017photo} represents a MSE loss of the 29th layer activations from a pre-trained VGG16 network $\phi_{vgg}$ to quantify the content difference.  $\lambda_{cont}$ is a hyper-parameter used to balance the L1 loss and content loss. Except that, We also utilize an L1 loss to achieve cycle consistence of $G_{T_1 \rightarrow T_2}$ and $G_{T_2 \rightarrow T_1}$:
\begin{small}
\begin{equation}
L_{cyc}^{T_1} = L_{l_1}(G_{T_2 \rightarrow T_1}(G_{T_1 \rightarrow T_2}(I_{T_1})), I_{T_1})
\label{eq5}
\end{equation}
\end{small}
Similarly, we apply $L_{gen}^{T_1}$ and $L_{cyc}^{T_2}$ to achieve the decoupling of $I_{T_2}$ style and content. The overall loss function in Stage 1 is shown as follows:
\begin{small}
\begin{equation}
L_{stage_1} = L_{gen}^{T_2} + L_{cyc}^{T_1} + L_{gen}^{T_1} + L_{cyc}^{T_2}
\label{eq6}
\end{equation}
\end{small}

\begin{figure}[t]
\centering
\includegraphics[width=1\linewidth]{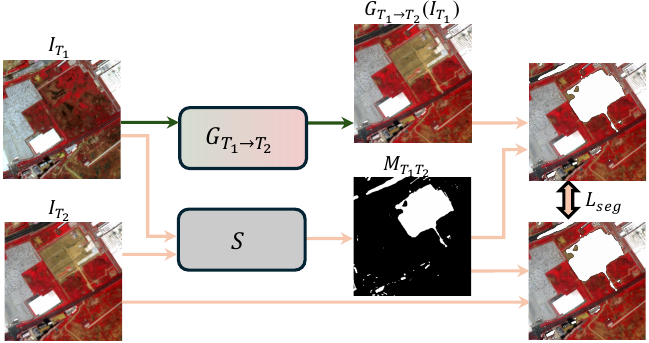} 
\caption{The design of the Change Segmentation module.
\label{CS}}
\end{figure}

\subsection{Change Segmentation module}

The CS module is utilized to capture the unreconstructed regions. The design of the CS module is shown in Fig.~\ref{CS}.
For paired patches $I_{T_1}$ and $I_{T_2}$, we first pass $I_{T_1}$ through the generator $G_{T_1 \rightarrow T_2}$ trained in CC module of Stage 1 to obtain $G_{T_1 \rightarrow T_2}(I_{T_1})$. We then input both $I_{T_1}$ and $I_{T_2}$ into a segmentation network $S$ to generate a mask map $M_{T_1T_2}$. The values in $M_{T_1T_2}$ range from 0 to 1, representing the probability of each pixel changing between the two paired image patches. Next, we apply an element-wise multiplication between $G_{T_1 \rightarrow T_2}(I_{T_1})$ and $(1 - M_{T_1T_2})$, as well as between $I_{T_2}$ and $(1 - M_{T_1T_2})$. The goal is to find a suitable $M_{T_1T_2}$ such that the difference between the two resulting masked images is minimized. Therefore, the purpose of the segmentation network is:
\begin{small}
\begin{equation}
\begin{split}
\begin{aligned}
L_{seg} = &L_{l_1}(G_{T_1 \rightarrow T_2}(I_{T_1}) \odot (1-M_{T_1T_2}), I_{T_2}\odot (1-M_{T_1T_2}))\\
+ &\lambda_{cont} \cdot L_{cont}(G_{T_1 \rightarrow T_2}(I_{T_1}) \odot (1-M_{T_1T_2}), I_{T_2}\odot (1-M_{T_1T_2}))\\
+ &\lambda_{reg} \cdot L_{reg}(M_{T_1T_2})
\end{aligned}
\end{split}
\label{eq7}
\end{equation}
\end{small}
where $M_{T_1T_2} = S(I_{T_1}, I_{T_2})$. $\odot$ represents element-wise multiplication and $\lambda_{reg}$ is a hyperparameter. The first two terms of $L_{seg}$ are analogous to $L_{gen}^{T_2}$, comprising an L1 loss and a content loss. Additionally, we incorporate a regularization loss $L_{reg}$ to encourage sparse mask areas. The $L_{reg}$ term is expressed as:
\begin{small}
\begin{equation}
L_{reg} = mean(M_{T_1T_2})
\label{eq8}
\end{equation}
\end{small}

\begin{figure}[t] %
\centering
\includegraphics[width=0.99\linewidth]{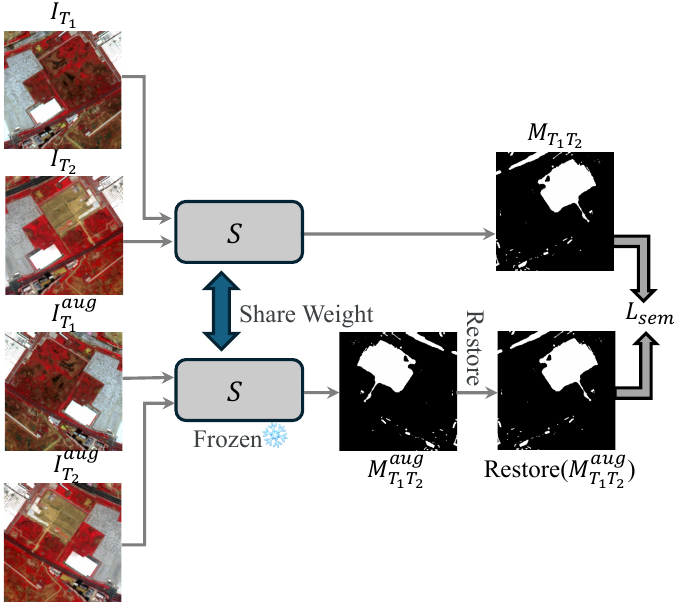} 
\caption{The architecture of the Semantic Consistency module.
\label{SC}}
\end{figure}

\subsection{Semantic Consistency module}

To ensure semantic consistency, we design the SC module, as shown in Fig.~\ref{SC}. Intuitively, horizontally flipping $I_{T_1}$ and $I_{T_2}$ (denoted as $I_{T_1}^{aug}$ and $I_{T_2}^{aug}$) and feeding them into the segmentation network $S$ should yield  $M_{T_1T_2}^{aug}$. Moreover, the result of horizontally flipping $M_{T_1T_2}^{aug}$, denoted as Restore($M_{T_1T_2}^{aug}$), should be equivalent to the original $M_{T_1T_2}$. To achieve this objective:
\begin{small}
\begin{equation}
L_{sem} = L_{l_1}({\rm Restore}(M_{T_1T_2}^{aug}), M_{T_1T_2})
\label{eq9}
\end{equation}
\end{small}
where $M_{T_1T_2}^{aug} = S(I_{T_1}^{aug}, I_{T_2}^{aug})$. The $\cdot^{aug}$ notation represents four augmentation options: identity, horizontal flip, vertical flip, and transpose. The Restore($\cdot$) operation refers to reverting the augmented inputs back to their original states. For the SC module, the gradients are not backpropagated through the branch corresponding to $I_{T_1}^{aug}$ and $I_{T_2}^{aug}$. Instead, the result of the augmented branch serves as a reference for the results of the $I_{T_1}$, $I_{T_2}$ branch.
Therefore, the full objective in Stage 2 is:
\begin{small}
\begin{equation}
L_{stage_2} = L_{seg} + \lambda_{sem} \cdot L_{sem}
\label{eq10}
\end{equation}
\end{small}
where $\lambda_{sem}$ is a hyperparameter. The third stage involves an iterative refinement process, where we alternately fine-tune the generative network $G_{T_1 \rightarrow T_2}$ and the segmentation network $S$, culminating in the improved overall performance of the change detection system. The objective function for this stage is defined as:
\begin{small}
\begin{equation}
L_{stage_3} = L_{gen}^{T_2} + L_{seg}
\label{eq11}
\end{equation}
\end{small}

\section{Experiments}

\subsection{Datasets}

\begin{figure}[t]
\centering
\includegraphics[width=0.99\linewidth]{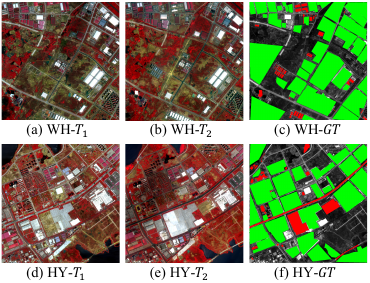} 
\caption{WH dataset and HY dataset, where changed areas are in red and unchanged regions are in green.
\label{dataset}}
\end{figure}

We utilize two high-resolution remote sensing change detection datasets to evaluate our proposed method. Each dataset comprises a bi-temporal image pair. The dimensions of each remote sensing image are 1000 $\times$ 1000 pixels, with spectral bands covering the red, green, blue, and near-infrared. The spatial resolution of these images is 4 meters per pixel.
WH dataset \cite{chen2019change} was acquired by the GF-2 sensor on April 4, 2016 and September 1, 2016, covering the city of Wuhan, China. The ground truth annotations for changed and unchanged regions are visualized in Fig.~\ref{dataset} (top), where the changed area (in red) encompasses 20026 pixels, the unchanged area (in green) covers 484143 pixels, and the remaining pixels are undefined.
Furthermore, HY dataset \cite{chen2019change} was also obtained by the GF-2 sensor, focusing on the Hanyang area of Wuhan city. The ground truth for changed and unchanged regions is depicted in Fig.~\ref{dataset} (bottom), with the changed area (in red) containing 59051 pixels, the unchanged area (in green) comprising 416404 pixels, and the remaining pixels being undefined.

\begin{figure*}[t]
\centering
\includegraphics[width=1\linewidth]{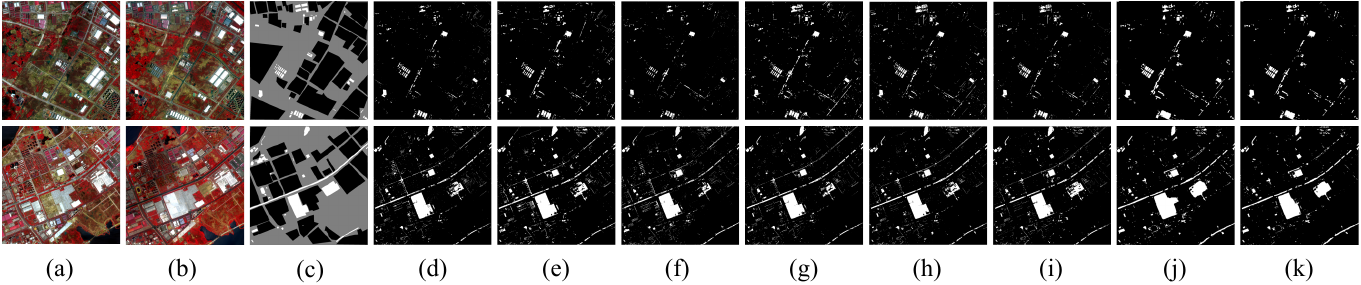} 
\caption{We compare our CCDF (k) against state-of-the-art methods, including (d) SVM-CD, (e) RNN-CD, (f) DSCN, (g) SiamCRNN(TR), (h) SiamCRNN(GRU), (i) SiamCRNN(LSTM), and (j) FCD-GAN, on the WH (top) and HY (bottom) datasets. (a) T1, (b) T2, and (c) ground truth are also provided. In (c-k), white denotes changes and black denotes no changes. Gray in (c) represents undefined areas.
\label{vis}}
\end{figure*}

\begin{table*}[t]
\caption{Comparison with other state-of-the-art approaches. The best result is in bold and the second best is underlined. All scores are reported in percentages.}
\renewcommand\arraystretch{1.1}
\begin{center}
\begin{tabular}{l|ccccccc|ccccccc}
\hline
\multirow{2}{*}{\diagbox [width=7em] {Methods} {Datasets}}& 
\multicolumn{7}{c|}{WH dataset}&\multicolumn{7}{c}{HY dataset}\cr\cline{2-15}
& OA & KC & Pre. & Rec. & F1 & mIOU & cIOU & OA & KC & Pre. & Rec. & F1 & mIOU & cIOU \cr\hline
\hline
SVM-CD \cite{hearst1998support}        
              & 97.55 & 63.30 & 75.96 & 56.10 & 64.54 & - & - &
                94.36 & 72.10 & 82.78 & 68.98 & 73.48 & - & - \\
RNN-CD \cite{lyu2016learning} 
              & 97.47 & 65.17 & 70.06 & 63.25 & 66.48 & 73.60 & 49.79 &
                94.38 & 72.96 & 80.51 & 72.22 & 76.14 & 77.65 & 61.47\\
DSCN \cite{zhan2017change} 
              & 97.32 & 55.82 & 78.55 & 44.84 & 57.09 & 68.61 & 39.95 & 
                93.40 & 67.09 & 78.71 & 64.28 & 70.77 & 73.79 & 54.76\\
SiamCRNN(TR) \cite{chen2019change}  
              & 97.67 & 69.19 & 70.98 & 69.83 & 70.40 & 75.96 & 54.33 &
                94.54 & 72.54 & 85.00 & 68.03 & 75.57 & 77.39 & 60.74\\
SiamCRNN(GRU) \cite{chen2019change} 
              & 97.78 & 70.21 & 73.33 & 69.49 & 71.36 & 76.60 & 55.47 &
                95.08 & 75.33 & 87.45 & 70.51 & 78.07 & 79.32 & 64.03\\
SiamCRNN(LSTM) \cite{chen2019change} 
              & 98.13 & 72.87 & 83.01 & 66.48 & 73.83 & 78.30 & 58.52 &
                95.39 & 77.01 & \underline{88.50} & 72.28 & 79.57 & 80.51 & 66.08\\
FCD-GAN \cite{wu2023fully} 
              & \underline{98.35} & \underline{76.71} & {\bf84.13} & 71.97 & 77.56 & \underline{80.83} & 63.36 &
                \underline{96.15} & \underline{81.90} & 86.60 & \underline{81.77} & \underline{84.09} & \underline{84.13} & \underline{72.55}\\
FCD-GAN$^1$ \cite{wu2023fully} 
              & 98.27 & 76.70 & 80.28 & \underline{75.30} & \underline{77.60} & 80.81 & \underline{63.40} & 
                95.92 & 80.99 & 84.97 & {\bf81.87} & 83.31 & 83.43 & 71.41\\
\hline
Ours 
              & {\bf98.42} & {\bf78.33} & \underline{83.23} & {\bf75.48} & {\bf79.15} & {\bf81.93} & {\bf65.49} &
                {\bf96.63} & {\bf83.81} & {\bf90.59} & 81.35 & {\bf85.71} & {\bf85.62} & {\bf75.00}\\
\hline
\end{tabular}
\label{main_tab}
\end{center}
\end{table*}

\begin{table*}[t]
\renewcommand\arraystretch{1.1}
\begin{center}
\caption{Ablation studies on HY dataset.}
\label{ablation}
\begin{tabular}{cccc|ccccccc}
\hline
$G_{T_1 \rightarrow T_2}$ & $G_{T_2 \rightarrow T_1}$ & CS module & SC module & OA & KC & Pre. & Rec. & F1 & mIOU & cIOU\\
\hline
\checkmark &   & \checkmark & & 
            95.92 & 80.99 & 84.97 & 81.87 & 83.31 & 83.43 & 71.41\\
\checkmark & \checkmark & \checkmark & & 
            96.37 & 82.65 & 89.03 & 80.82 & 84.70 & 84.71 & 73.46\\
\checkmark&  & \checkmark &\checkmark& 
            96.35 & 82.53 & 88.88 & 80.76 & 84.60 & 84.63 & 73.32\\
\checkmark & \checkmark & \checkmark &\checkmark & 
            96.63 & 83.81 & 90.59 & 81.35 & 85.71 & 85.62 & 75.00\\
\hline
\end{tabular}
\end{center}
\end{table*}

\subsection{Implementation Details}

We implement our method based on PyTorch \cite{paszke2017automatic} and conduct the training process on an NVIDIA GeForce RTX 4090 GPU, utilizing a batch size of 10. For the backbone architecture, we employ generator networks from Wu et al.\cite{wu2023fully} and a segmentation network from Daudt et al.\cite{daudt2018fully} and Hou et al. \cite{hou2019w}.
The input remote sensing images from the datasets have a spatial dimension of $W \times H \times C$, where $W$ = 1000, $H$ = 1000, and $C$ = 4. Due to the large image size, we cannot process the entire image directly. Instead, we subdivide the large images into smaller patches with a size of 224 $\times$ 224 $\times$ 4, maintaining an overlap of 12 pixels between neighboring patches.
For the training process, we utilize an Adam optimizer with a warmup strategy. In Stage 1 and Stage 2, the minimum and maximum learning rates are set to 1e-5 and 3e-4, respectively. In Stage 3, we fine-tune the generator and segmentation networks using a warmup strategy with a minimum learning rate of 1e-5 and a maximum learning rate of 1e-4. The training epochs for Stage 1, Stage 2, and Stage 3 are 30, 30, and 50, respectively. We use a threshold of 0.5 to determine the change/non-change regions in the segmentation output.
For the hyperparameters, we choose $\lambda_{cont}$ = 0.2, $\lambda_{reg}$ = 0.75, and $\lambda_{sem}$ = 0.7 for WH dataset, and $\lambda_{cont}$ = 0.4, $\lambda_{reg}$ = 0.65, and $\lambda_{sem}$ = 0.7 for HY dataset. We build our experimental environment based on the open-source FCD-GAN \cite{wu2023fully} code and compare our method with the re-implemented FCD-GAN\footnote{The re-implemented FCD-GAN based on the release code \cite{wu2023fully_code}.} \cite{wu2023fully}.

\subsection{Evaluation Metrics}

Our approach is evaluated using seven metrics, including overall accuracy (OA), kappa coefficient (KC), precision rate (Pre.), recall rate (Rec.), F1-score (F1), mean Intersection-over-Union (mIOU), and changed Intersection-over-Union (cIOU).

\subsection{Performance Comparisons}

We assess the performance of our proposed Consistency Change Detection Framework (CCDF) and benchmark it against several state-of-the-art approaches, including SVM-CD \cite{hearst1998support}, DSCN \cite{zhan2017change}, RNN-CD \cite{lyu2016learning}, SiamCRNN \cite{chen2019change}, and FCD-GAN \cite{wu2023fully}. To ensure statistical robustness, each experiment is repeated five times, and the reported results represent the average of these five trials.

The results on  WH and HY datasets are shown on the left and right of Table \ref{main_tab} respectively. Our CCDF achieves F1 scores of 79.15\% (WH) and 85.71\% (HY), as well as cIOU of 65.49\% (WH) and 75.00\% (HY). Fig.~\ref{vis} shows results obtained by the proposed method and comparison methods on WH dataset and HY dataset. Our Cycle Consistency (CC) module helps to reduce overfitting in the generator-based reconstruction and is better for the subsequent segmentor. Moreover, the Semantic Consistency (SC) module ensures coherence between the original and augmented images. Together, these two parts enhance the change detection capabilities of our approach.

\subsection{Ablation Studies}

We conduct ablation experiments on HY dataset. Our method consists of three modules: a CC module (containing $G_{T_1 \rightarrow T_2}$ and $G_{T_2 \rightarrow T_1}$), a SC module, and a Change Segmentation (CS) module. The baseline method is the FCD-GAN ($G_{T_1 \rightarrow T_2}$ + CS module). 
We evaluate the improvement made by the CC module. We test the results with and without the SC module. Without the SC module, comparing the 1st row and the 2nd row of Table \ref{ablation}, the CC module brings about a 1.7\% and  2.9\% performance boost over the baseline FCD-GAN in terms of F1 and cIOU. With the SC module, the results are shown in the 3rd and 4th rows. The CC module brings 1.3\% and 2.3\% performance boosts. Our CC module helps $G_{T_1 \rightarrow T_2}$ retain more content and less style, leading to better results.
The contribution of the SC module is shown in the 2nd and 4th row of Table \ref{ablation}. The last row of Table \ref{ablation} shows that our CC module and SC module jointly contributed to the final result, which achieves 85.71 and 75.00 in F1 and cIOU, a 2.9\% and 5\% performance boost over the baseline.

\section{Conclusion}

In this paper, we propose a Consistency Change Detection Framework (CCDF) to tackle the unsupervised remote sensing change detection problem. Our CCDF employs a Cycle Consistency (CC) module to construct two global style transfer generators, which helps reduce the loss of content and mitigate generator overfitting. Furthermore, the Semantic Consistency (SC) module, guarantees detailed reconstruction. Extensive experiments validate the superior performance of our method, as well as the eﬀectiveness of each component.

\section*{Acknowledgment}

The authors gratefully acknowledge the support of the National Key Research and Development Program of China (2023YFA1608100), the JC STEM Lab of AI for Science and Engineering, the Hong Kong Jockey Club Charities Trust, the Research Grants Council of Hong Kong (Project No. CUHK14213224), the National Natural Science Foundation of China (NSFC, Grant Nos. 12173037, 12233008), the CAS Project for Young Scientists in Basic Research (No. YSBR-092), and Cyrus Chun Ying Tang Foundations.

\bibliographystyle{IEEEbib}
\bibliography{icme2025references}

\end{document}